\documentclass[letterpaper, 10 pt, conference]{ieeeconf}  

\IEEEoverridecommandlockouts                              

\usepackage{bm}
\usepackage{amsmath} 
\usepackage{amssymb}  
\usepackage{comment}
\usepackage{algorithm}
\usepackage{algpseudocode}
\usepackage{booktabs}
\usepackage{multirow}
\usepackage{graphicx}
\usepackage{tabularx}
\usepackage{subcaption}
\usepackage{hyperref} 

\usepackage{xcolor}
\usepackage[%
  backend=biber,
  url=true,
  style=ieee, 
  sorting=none, 
  maxnames=4,
  minnames=3,
  maxbibnames=99,
  giveninits,
  uniquename=init]{biblatex} 

\newcommand{\todo}[1]{{\leavevmode\color{red}[TODO: #1]}}

\title{\LARGE \bf
End-to-end QP-based policies:\\ A unified perspective on robust control and robot learning }

\author{
Fausto Vega$^{1}$,
Priyanka Supraja Balaji$^{2}$,
Chase Dunaway$^{1}$,
Joe Koszut$^{1}$\\
Jon Arrizabalaga$^{1}$,
and Zachary Manchester$^{1}$
\thanks{$^{1}$Authors are with the Department of Aeronautics and Astronautics,
Massachusetts Institute of Technology, Cambridge, MA, USA {\tt\small \{vegaf, chasead7,jkoszut,jonarri,zacm\}@mit.edu}}
\thanks{$^{2}$Author is with the Department of Mechanical Engineering,
California Institute of Technology, Pasadena, CA, USA {\tt\small psupraja@caltech.edu}}
}

\bibliography{references}

\begin{document}

\maketitle
\thispagestyle{empty}
\pagestyle{empty}

\begin{abstract}
We present an end-to-end QP-based policy framework that enables systematic policy construction with minimal domain-specific design, while preserving the transparency and interpretability of model-based control. The proposed policy representation supports both domain-randomized model-based auto-tuning, where policy parameters are optimized over distributions of disturbances and model variations, and black-box policy construction, where the problem is formulated in terms of a (possibly) unknown model without requiring explicit notions of states, inputs, or the underlying system dynamics. We establish connections to existing policy representations and control paradigms, including robust control, multilayer perceptrons, and robot learning, and interpret our end-to-end QP policies as a common abstraction of these approaches. We validate the resulting framework in both simulation and hardware, demonstrating a broad range of applications spanning robustness, automatic policy tuning, and control under unknown system dynamics. 

\end{abstract}



\begin{flushleft}
\textbf{Website}: {\footnotesize
\url{https://roboticexplorationlab.org/e2eqp/}}
\end{flushleft}
\section{INTRODUCTION}

Constrained optimization sits at the very core of robotics. It appears across virtually all its disciplines, including planning, control, state estimation, simulation, and many others. The early acrobatic maneuvers of Atlas and the rocket landings of SpaceX are a testament to the fundamental role of convex optimization in some of the field's most remarkable achievements~\cite{kuindersma2014efficiently,blackmore2016autonomous}.

Despite these achievements, the role of convex optimization within robotics has changed substantially in recent years. This shift reflects a transformation of the robotics landscape: modular architectures are giving way to unified end-to-end pipelines, while analytical models are complemented by---or replaced with---black-box representations, such as simulators or world models, whose inner state and derivatives may be unavailable to the user~\cite{todorov2012mujoco, hou2026world}. In this new setting, convex optimization is no longer used only to compute optimal solutions efficiently. Increasingly, the optimization problem itself becomes part of what must be adapted to the task and the data at hand.

A key enabler of this transition has been differentiability~\cite{krantz2002implicit}. Recent methods not only solve convex optimization problems efficiently, but also differentiate their solutions with respect to the problem data~\cite{amos2017optnet, agrawal2019differentiable, tracy2024differentiability, bravo2026turbompc, arrizabalaga2026differentiable, holmes2025sdprlayers}. This makes the parameters of an optimization problem quantities that can be tuned through downstream gradient descent, bringing optimization into the end-to-end learning loop.

Differentiable optimization has fueled a rapidly growing interface between optimization and learning. The impact has been particularly visible in low-level control, where model-predictive control (MPC) is increasingly treated not as a fixed controller, but as a component that can be adapted from data. Existing approaches range from directly differentiating through MPC to tune its costs and constraints~\cite{adabag2026differentiable}, to augmenting it with learned residuals~\cite{salzmann2023real}, and integrating it within reinforcement learning (RL) architectures~\cite{romero2024actor, reiter2026synthesis}. Similar ideas have been extended to differentiable control barrier functions~\cite{xiao2023barriernet, morton2025safe} and learned optimization methods tailored to specific problem distributions~\cite{sambharya2026learning}. 

However, in most of the approaches cited above, optimization remains one component within a broader architecture: a differentiable layer, an MPC controller with learned parameters, or a mechanism for imposing structure and constraints. In parallel, work has sought deeper connections between model-based and data-driven control from theoretical and behavioral perspectives~\cite{markovsky2021behavioral}. Here, we pursue a complementary question: rather than combining separately designed model-based and learning components, \emph{can the optimization problem itself serve as the policy representation? And can this representation expose common structure across control and learning?}

We show that it can. We formulate the optimization problem itself as the policy representation, with its defining parameters learned end-to-end from downstream objectives. This formulation accommodates both model-based control, where analytical structure is retained and tuned over distributions of models and disturbances, and black-box control, where policies are learned directly from input-output behavior without explicit notions of states, inputs, or dynamics. It also exposes common structure across seemingly different paradigms, including robust control, multilayer perceptrons, and domain-randomized robot learning. In this work, we instantiate this perspective using Quadratic Programs (QPs), while the underlying ideas extend naturally to more general conic convex formulations~\cite{agrawal2019differentiating}. 

In particular, our contributions are:

\begin{enumerate}

    \item An end-to-end differentiable QP-based policy representation that enables systematic policy construction with minimal domain-specific design, while preserving the transparency and interpretability of model-based control.

    \item Connections between our QP-based policy representation and established paradigms in control and learning, including robust control ($H_2$ and $H_\infty$), machine learning (multilayer perceptrons), and robot learning (domain-randomized reinforcement learning).

    \item A broad portfolio of simulation and real-world experiments showcasing the aforementioned connections, as well as the ease of use and deployment of the proposed framework.

\end{enumerate}

The remainder of the paper is organized as follows: Section~\ref{sec:e2eqp} introduces the end-to-end QP-based policy representation, then Section~\ref{sec:method} establishes connections to the aforementioned control and learning paradigms. Section~\ref{sec:experiments} presents a set of simulation and hardware experiments to demonstrate our results on practical robotics problems. Finally, Section~\ref{sec:conclusions} summarizes our conclusions  .

\begin{figure*}[t]
\centering

\begin{subfigure}[t]{0.32\textwidth}
    \centering
    \includegraphics[width=\linewidth]
    {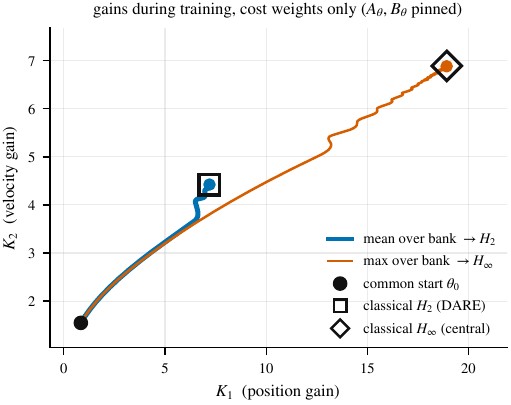}
    \caption{Recovery of the feedback gain $K_\theta$ under
    $H_2$- and $H_\infty$-type objectives.}
    \label{fig:h2-hinf-recovery}
\end{subfigure}
\hfill
\begin{subfigure}[t]{0.32\textwidth}
    \centering
    \includegraphics[width=\linewidth]
    {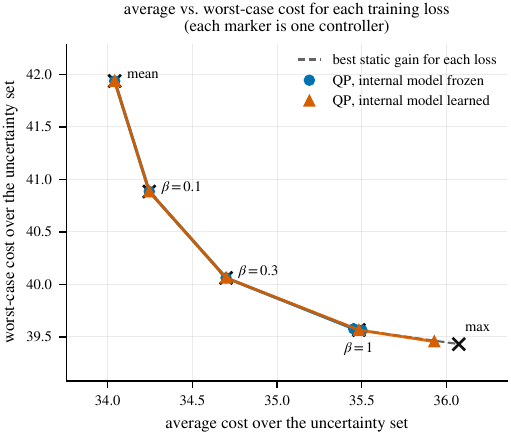}
    \caption{Tradeoff between average- and worst-case performance
    under entropic risk over model uncertainty.}
    \label{fig:pareto-risk}
\end{subfigure}
\hfill
\begin{subfigure}[t]{0.32\textwidth}
    \centering
    \includegraphics[width=\linewidth]
    {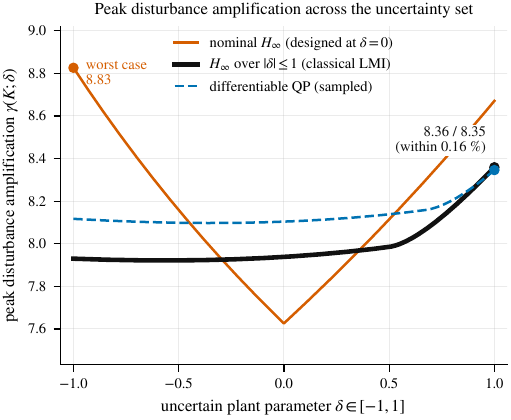}
    \caption{Worst-case disturbance amplification across the
    uncertain plant family}
    \label{fig:robust-third}
\end{subfigure}

\caption{Robust-control connections of the proposed QP policy on a double integrator regulation task.
(a) Disturbance uncertainty: optimizing the outer objective approximates
$H_2$- and $H_\infty$-type feedback gains.
(b) Model uncertainty: entropic risk interpolates between average- and
worst-case performance.
(c) Combined model and disturbance uncertainty: the learned QP policy
is compared with nominal $H_\infty$ and worst case $H_\infty$ under plant uncertainty.}
\label{fig:robust-control-results}
\end{figure*}

\section{END-TO-END QP BASED POLICIES}
\label{sec:e2eqp}


We consider policies whose control action is obtained from the solution of a parameterized quadratic program. Rather than fixing a single QP structure, we allow the formulation to vary depending on the problem. In this work, we consider formulations ranging from a projection QP to a trajectory-optimization QP with predictive dynamics, horizon coupling, costs, and constraints. These different structures induce different classes of control laws. 


Given the current state $x$, we define the QP policy through
\begin{equation}
\begin{aligned}
z^\star(x;\theta)
=
\arg\min_z \quad &
\frac{1}{2} z^\top H_\theta z
+
\left(F_\theta x + q_\theta\right)^\top z \\
\text{s.t.}\quad &
G_\theta z \le h_\theta, \\
&
E_\theta z = b_\theta,
\end{aligned}
\label{eq:generic_qp_policy}
\end{equation}
with the control action recovered as
\begin{equation}
u
=
\pi_\theta^{\mathrm{QP}}(x)
=
C_\theta z^\star(x;\theta) + d_\theta.
\label{eq:qp_readout}
\end{equation}
Here, $z$ represents the QP decision variable and $\theta$ collects the learnable parameters defining the objective, constraints, and the control from the QP solution.

For a given QP formulation, the parameters $\theta$ determine a particular control law within the QP policy class. We optimize these parameters directly from closed-loop task performance. Consider a plant with dynamics
\begin{equation}
x_{t+1}
=
f_p(x_t,u_t,w_t),
\label{eq:true_plant}
\end{equation}
where $p$ represents the plant parameters and $w_t$ is an exogenous disturbance. Applying the QP policy recursively yields
\begin{equation}
x_{t+1}
=
f_p\!\left(
x_t,
\pi_\theta^{\mathrm{QP}}(x_t),
w_t
\right).
\label{eq:closed_loop_dynamics}
\end{equation}
For an initial condition $x_0$, plant instance $p$, and disturbance sequence $w=\{w_t\}_{t=0}^{T-1}$, the resulting closed-loop trajectory is evaluated through the task objective
\begin{equation}
J(\theta;x_0,p,w)
=
\sum_{t=0}^{T-1}
\ell(x_t,u_t)
+
\ell_f(x_T),
\label{eq:trajectory_loss}
\end{equation}
where $T$ is the simulation horizon, $\ell$ is the stage loss, and $\ell_f$ is the terminal loss. 

Training is performed over a batch of $M$ closed-loop rollouts. One possible choice of outer objective is the empirical mean of the rollout objectives,
\begin{equation}
\mathcal{L}_{\mathrm{mean}}(\theta)
=
\frac{1}{M}
\sum_{i=1}^{M}
J\!\left(
\theta;
x_0^{(i)},
p^{(i)},
w^{(i)}
\right).
\label{eq:batch_objective}
\end{equation}
The sampled quantities can be varied according to the desired training objective. Sampling $x_0^{(i)}$ trains the policy over a distribution of initial conditions, sampling $p^{(i)}$ introduces domain randomization over plant uncertainty, and sampling $w^{(i)}$ exposes the policy to exogenous disturbances. These sources of variation can be considered independently or jointly.

The differentiability of the QP and plant rollout allows gradients of the outer objective to be propagated through all the closed-loop trajectories in the batch. Each rollout $i$ produces an individual trajectory  loss $J^{(i)}$, which is then combined through the chosen outer objective $\mathcal{L}$. The resulting computation graph can be summarized as
\begin{equation}
\theta
\longrightarrow;
{z_t^{\star(i)}}
\longrightarrow;
{u_t^{(i)}}
\longrightarrow;
{x_{t+1}^{(i)}}
\longrightarrow;
{J^{(i)}}
\longrightarrow;
\mathcal{L},
\label{eq}
\end{equation}
where $i=1,\ldots,M$ indexes the rollouts and $t=0,\ldots,T-1$ indexes time. This enables direct optimization of the policy parameters using gradient-based methods.

The outer objective applied to the rollout losses can be chosen independently of the underlying QP structure. Replacing the empirical mean in \eqref{eq:batch_objective} with a risk-sensitive or worst-case operator changes the criterion used to select the control law while leaving the QP structure unchanged. The separation between QP structure and outer learning objective is central to our framework: the QP specifies which policies can be represented, while the outer objective determines which policy is learned. We use this distinction in the next section to establish connections with neural-network policies and robust-control objectives. 

\section{QP POLICIES ACROSS CONTROL AND LEARNING}
\label{sec:method}

\subsection{Equivalence to Single-Hidden-Layer Neural Networks}


The projection QP provides a direct connection between QP-based policies and neural-network policies. In particular, the ReLU activation can be represented exactly as the Euclidean projection onto the nonnegative orthant. Consider a single-hidden-layer ReLU policy
\begin{equation}
\pi_{\mathrm{ReLU}}(x)
=
W_2\,\mathrm{ReLU}(W_1x+b_1)+b_2,
\label{eq:relu_policy}
\end{equation}
where $W_1$, $W_2$, $b_1$, and $b_2$ are the network parameters. The hidden activation can equivalently be obtained as the solution of the projection QP
\begin{equation}
z^\star(x)
=
\arg\min_{z \geq 0}
\frac{1}{2}
\left\|
z-(W_1x+b_1)
\right\|_2^2.
\label{eq:relu_qp}
\end{equation}
The solution of \eqref{eq:relu_qp} is

\begin{equation}
z^\star(x)
=
\mathrm{ReLU}(W_1x+b_1),
\label{eq:relu_qp_solution}
\end{equation}
so the corresponding QP policy satisfies
\begin{equation}
\pi_{\mathrm{QP}}(x)
=
W_2 z^\star(x)+b_2
=
\pi_{\mathrm{ReLU}}(x).
\label{eq:relu_qp_equivalence}
\end{equation}

Thus, a single-hidden-layer ReLU policy is a special case of the QP
policy formulation introduced in
\eqref{eq:generic_qp_policy}--\eqref{eq:qp_readout}
~\cite{amos2019differentiable}.
The projection QP and the corresponding single-hidden-layer ReLU network therefore have the same expressive capacity under this parameterization. This equivalence also provides a direct demonstration of end-to-end QP learning of a piecewise-affine control law, which we illustrate on a nonlinear cartpole swing-up task in Section  \ref{sec:experiments}.





\subsection{Recovering Robust Control Objectives}

We next connect the proposed framework to classical robust-control objectives in the special case of an unconstrained linear time-invariant (LTI) system. We do so by isolating the effect of the outer objective from the plant and QP structure. To facilitate understanding, we accompany the discussion throughout this section with an illustrative double-integrator example.


Consider a batch containing plant instances $p^{(i)}$ and disturbance realizations $w^{(j)}$. Each closed-loop rollout produces a loss
\begin{equation}
J_{ij}(\theta)
=
J\!\left(
\theta;
p^{(i)},
w^{(j)}
\right),
\end{equation}
which we combine through the nested objective
\begin{equation}
\mathcal{L}(\theta)
=
\rho_p
\left(
\left\{
\rho_w
\left(
\left\{
J_{ij}(\theta)
\right\}_{j=1}^{M}
\right)
\right\}_{i=1}^{N}
\right).
\label{eq:two_knob}
\end{equation}
Here, $\rho_w$ determines how performance is aggregated over exogenous disturbances, while $\rho_p$ determines how performance is aggregated over model uncertainty. The QP policy is then optimized according to
\begin{equation}
\theta^\star
=
\arg\min_\theta
\mathcal{L}(\theta).
\end{equation}

For this analysis, we use the trajectory-optimization QP
\begin{equation}
\begin{aligned}
\min_{\hat{x}_{0:H},\,u_{0:H-1}}
\quad &
\sum_{t=0}^{H-1}
\left(
\hat{x}_t^\top Q_\theta \hat{x}_t
+
u_t^\top R_\theta u_t
\right)
+
\hat{x}_H^\top P_\theta \hat{x}_H
\\
\text{s.t.}\quad
&
\hat{x}_0 = x,
\\
&
\hat{x}_{t+1}
=
A_\theta \hat{x}_t
+
B_\theta u_t .
\end{aligned}
\label{eq:robust_qp_policy}
\end{equation}
In the unconstrained LTI case, the first control action reduces to a linear state-feedback law
\begin{equation}
\pi_\theta^{\mathrm{QP}}(x)
=
-K_\theta x.
\end{equation}
Thus, learning the QP parameters $\theta$ indirectly selects a feedback gain $K_\theta$ from the class induced by the QP parameterization.

We first consider disturbance uncertainty for a fixed plant
\begin{equation}
x_{k+1}
=
A x_k + B u_k + B_w w_k,
\end{equation}
with performance signal
\begin{equation}
e_k
=
\begin{bmatrix}
Q_e^{1/2}x_k\\
R_e^{1/2}u_k
\end{bmatrix}.
\end{equation}
We evaluate the closed-loop response to a bank of sinusoidal disturbances spanning the discrete-time frequency range. For each frequency $\omega_j$, the squared frequency-response gain is approximated by
\begin{equation}
G_j(\theta)
=
\frac{
\sum_k \|e_k^{(j)}\|_2^2
}{
\sum_k \|w_k^{(j)}\|_2^2
}.
\label{eq:Gj}
\end{equation}
%
A weighted average of these gains over frequency gives an $H_2$-type objective \cite{megretski20006},
\begin{equation}
\mathcal{L}_2(\theta)
=
\sum_{j=0}^{M-1}
\alpha_j G_j(\theta),
\end{equation}
where $\alpha_j$ are normalized trapezoidal quadrature weights associated
with the uniformly sampled frequencies $\omega_j$. 
For the $H_\infty$-type objective, we instead select the largest gain
across the sampled frequencies,
\begin{equation}
\mathcal{L}_\infty(\theta)
=
\max_{0\leq j\leq M-1} G_j(\theta).
\end{equation}

%
As the frequency discretization is refined, these objectives approximate the squared $H_2$ and $H_\infty$ norms of the closed-loop disturbance-to-performance map. Fig.~\ref{fig:h2-hinf-recovery} shows that optimizing the QP parameters under these two outer objectives approximates the corresponding classical feedback gains \cite{basar1989dynamic}, \cite{dietz2008analysis} \cite{zhou1996robust}. 

We next isolate uncertainty in the plant. Let $J_i(\theta)$ denote the closed-loop trajectory cost obtained on plant instance $p^{(i)}$. The operator $\rho_p$ may then be chosen as the empirical mean, the maximum, or an intermediate risk-sensitive objective. In particular, we use the entropic risk \cite{guth2025approximation}
\begin{equation}
\mathcal{L}_{\beta}(\theta)
=
\frac{1}{\beta}
\log
\left(
\frac{1}{N}
\sum_{i=1}^{N}
\exp\!\left(\beta J_i(\theta)\right)
\right).
\label{eq:k2_entropic}
\end{equation}
As $\beta\rightarrow 0$, this objective approaches the mean plant cost, whereas increasing $\beta$ progressively emphasizes high-cost plants and approaches the worst case as $\beta\rightarrow\infty$. Hence, $\beta$ provides a continuous interpolation between average-case domain randomization and worst-case robust design. This behavior is illustrated in Fig.~\ref{fig:pareto-risk}.

Finally, both uncertainty sources can be activated simultaneously. Choosing
\begin{equation}
\rho_w = \max_j,
\qquad
\rho_p = \max_i,
\end{equation}
gives
\begin{equation}
\mathcal{L}_{\mathrm{rob}}(\theta)
=
\max_i \max_j G_{ij}(\theta),
\label{eq:rc_loss}
\end{equation}
which is a sampled approximation of the worst-case $H_\infty$ performance over an uncertain plant family. We compare this learned QP policy with $H_\infty$ design based on a common-Lyapunov-function LMI \cite{boyd1994linear}. The classical formulation minimizes a certified upper bound on worst-case disturbance amplification, whereas \eqref{eq:rc_loss} directly minimizes its sampled value. 
The resulting worst-case performance over the uncertainty set is shown in Fig.~\ref{fig:robust-third}.

The examples in Fig. \ref{fig:robust-control-results} demonstrate that, under the restrictive unconstrained LTI setting, the proposed outer-learning formulation recovers several familiar robust-control objectives through the choice of $\rho_w$ and $\rho_p$. Importantly, the QP policy representation itself is unchanged: the QP defines the class of admissible controllers, while the outer objective determines the desired notion of robustness. In Section~\ref{sec:experiments}, we use this same separation beyond the LTI setting to train constrained QP policies directly through nonlinear and uncertain robotic systems.

\begin{figure}[t]
    \centering
    \includegraphics[width=\columnwidth]{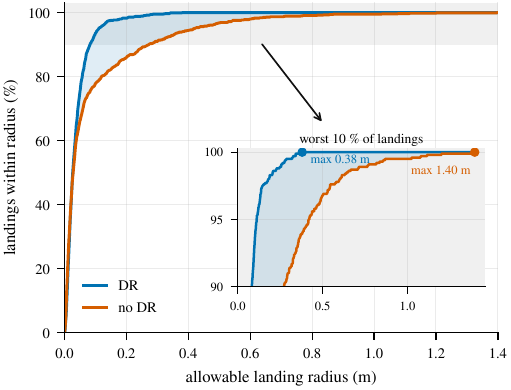}
    \caption{Percentage of landings within a specified landing radius for
    1000 vehicles, with and without domain randomization (DR). The inset
    highlights the upper landing-error tail.}
    \label{fig:rocket_landing_success}
\end{figure}

\subsection{Data-Driven Predictive Control}





The QP-policy abstraction also includes data-driven predictive controllers,
where future trajectories are represented directly from measured
input-output data rather than through an explicit state-space model
\cite{willems2005note,coulson2019data}. Originally developed for LTI systems, these behavioral approaches have also been extended to nonlinear settings through local, lifted, and kernel-based representations \cite{lian2021koopman}, \cite{naf2025choose}, \cite{huang2023robust}.

This further illustrates the role of QP structure in defining the resulting
feedback policy. Model-based predictive control constrains future
trajectories through an explicit dynamics model, data-driven predictive
control constrains them through measured input-output data, and the
projection QP introduced previously represents a piecewise-affine feedback
policy. More generally, consistent with the end-to-end QP-policy formulation introduced in Section~\ref{sec:e2eqp}, the QP structure determines the class of feedback laws that can be represented, while the outer closed-loop objective determines which policy within that class is learned.


More broadly, the same separation between QP structure and the outer objective also connects the QP-policy representation to robot learning. Once the QP structure specifies the policy class, its parameters can be optimized directly from closed-loop experience. In particular, sampling initial conditions, disturbances, and model parameters during training leads to domain-randomized policy optimization, which we consider in one of our experiments.




\section{EXPERIMENTS}
\label{sec:experiments}
We divide our experimental validation into three parts, each highlighting a different capability of the proposed QP-policy framework. We first study robustness under model mismatch by using domain randomization to tune a constrained MPC for a lunar lander with unmodeled slosh dynamics. We then examine the effect of QP structure on the resulting policy class through the projection-QP formulation, showing that a QP-based policy equivalent to a single-hidden-layer ReLU network can learn the nonlinear CartPole swing-up task. Finally, we evaluate the full end-to-end learning pipeline on hardware, including both state estimation and control, through deployment on a quadrotor.
\subsection{Robust Control with Domain Randomization}

To demonstrate the effect of domain randomization (DR) on control robustness, we consider a reusable lunar landing scenario. Reusability requires enough propellant to be carried not only for landing but also for ascent and return to orbit. Consequently, the lander must touch down with a large propellant mass fraction ($\approx 0.5$), resulting in non-negligible slosh dynamics. These dynamics are unobservable and thus cannot be included in a controller's internal model. Hence, we seek to automate the design of an MPC that explicitly respects actuator limits while gaining implicit robustness to structural model mismatch through domain randomization.

The lander is modeled as a planar rigid body with three degrees of freedom: vertical position $z$, horizontal position $y$, and orientation $\phi$. An additional coordinate $\psi$ is introduced to represent the first slosh mode, following the model of~\cite{howell2021direct}. The simulated plant used for training incorporates all four coordinates $(z, y, \phi, \psi)$ while the MPC is formulated only on the rigid subsystem $(z, y, \phi)$. The control inputs are thrust magnitude and gimbal angle.

Two controller variants are designed, sharing the same architecture, horizon, and constraint set. They differ only in the distribution used to train their running cost weights $Q$ and $R$ and terminal cost $P$. The baseline controller is trained with a fixed slosh pendulum model while the DR controller is trained over a distribution of slosh mass, pendulum length, and damping ratio. Each controller is then evaluated on the same fleet of 1000 vehicles exposed to a dispersion in initial conditions and slosh model parameters.

Fig.~\ref{fig:rocket_landing_success} reports the fraction of the vehicle fleet landing within a given distance from the landing target. The robustness of the DR controller is highlighted in the inset: its worst landing error is less than a third of the nominal controller's largest error. Slosh parameters are difficult to identify during flight and change across the course of a mission. Randomizing them during the controller design phase costs nothing at run time since the deployed controller retains the same structure as the nominal one, while avoiding the tail errors that come with a nominal controller tuned around a single design point. This shows that domain randomization over sensitive and unknown parameters like slosh in a lunar lander can provide considerable robustness over using a single fixed model. We next examine the effect of QP structure on the class of feedback laws that can be learned.

\begin{figure}[t]
    \centering
    \includegraphics[width=\columnwidth]{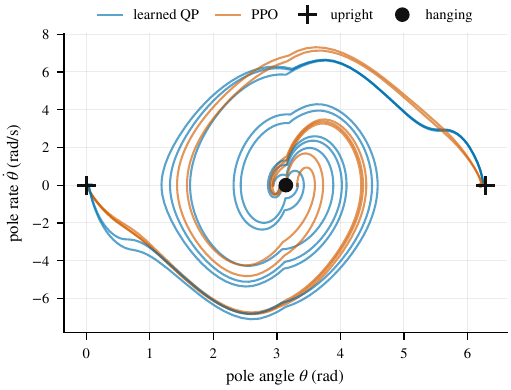}
    \caption{Closed-loop phase portraits of the learned projection-QP policy
and the single-hidden-layer ReLU policy from the same five initial
conditions near the hanging equilibrium. Both policies generate swing-up
trajectories and converge to the upright equilibrium.}
    \label{fig:cartpole-phase}
\end{figure}
\subsection{Single Hidden Layer ReLU Networks}
We evaluate the projection-QP policy on the nonlinear CartPole swing-up task to study the effect of QP structure on the resulting feedback law. The state is
$x=[p,\dot p,\theta,\dot\theta]^\top$, where $p$ and $\dot p$ describe
the cart motion and $\theta$ and $\dot\theta$ the pendulum motion. The
scalar input $u$ is the horizontal force applied to the cart.

This example highlights the role of QP structure in shaping the resulting
policy class. A linear-MPC QP relies on a local linear model, making it
well suited to regulation near the linearization point. The projection QP
in Section~\ref{sec:method} instead represents the same piecewise-affine
policy class as a single-hidden-layer ReLU network without relying on a
local predictive model.

We optimize
$\theta=\{W_1,b_1,W_2,b_2\}$ from closed-loop performance over batches of
$B=128$ differentiable CartPole rollouts from sampled initial conditions.
We train with short horizons and augment the objective with a local LQR
cost-to-go when the terminal state approaches the upright equilibrium.

Figure~\ref{fig:cartpole-phase} shows that the learned QP policy generates
the energy required for swing-up and then stabilizes the pendulum about
the upright equilibrium. Although each policy evaluation solves a convex
QP, the nonnegativity constraints on the hidden variables become active
in different regions of the state space, producing different affine
feedback laws. The resulting piecewise-affine policy is therefore able to
generate the nonlinear closed-loop behavior required for swing-up.

This experiment shows that a QP policy can represent behavior beyond local regulation, with its capabilities determined by the structure of the optimization problem. Using the projection formulation, the policy recovers the ReLU policy class and learns a controller that achieves nonlinear swing-up directly from closed-loop performance using the same end-to-end framework introduced in Section \ref{sec:e2eqp}. We next evaluate the framework on hardware, where state estimation and control are learned within the same end-to-end pipeline.

\begin{figure}[t]
    \centering
    \includegraphics[width=\columnwidth]{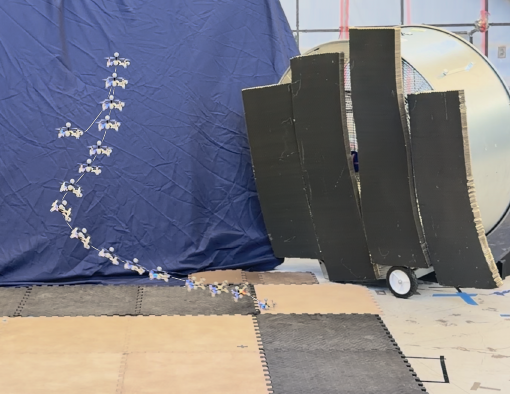}
    \caption{Our end-to-end QP policy applied to a Crazyflie drone under a $5\ \mathrm{m/s}$ crosswind generated by an industrial fan with a flow-conditioning rack. Estimator, offset-compensation, and cost parameters are learned end to end, removing manual tuning and outperforming the hand-tuned baselines.} 
    \label{fig:wind_setup}
\end{figure}

\subsection{End-to-end Learning in Hardware}

Finally, we showcase how our QP-based policy framework enables systematic policy construction with minimal domain-specific design. We evaluate the framework on hardware by incorporating state estimation and control into the differentiable closed-loop pipeline. Specifically, we consider position regulation of a Crazyflie quadrotor under a strong external wind disturbance, demonstrating end-to-end tuning of a structured QP policy together with a disturbance observer.

The experimental setup is shown in Fig. \ref{fig:wind_setup}. A Crazyflie quadrotor flies in a motion-capture environment while exposed to a $5 \ \mathrm{m/s}$ crosswind generated by an industrial fan\footnote{The industrial fan is Venco MAC-48-316-C7-J1 (560W, 19500CFM)} operating at full speed. A flow-conditioning rack is used to produce a more repeatable wind field. The Crazyflie begins approximately $0.5 \ \mathrm{m}$ from the fan and flies to a target position $1.9 \ \mathrm{m}$ downstream, where it maintains its position under the wind disturbance. 

To account for the persistent wind disturbance, we augment the state with an estimated disturbance $c$ and jointly estimate the state and disturbance using a Luenberger observer~\cite{luenberger1966observers,pannocchia2015offset}:
\begin{equation}
    \begin{bmatrix}
    x_{k+1} \\
    c_{k+1}
    \end{bmatrix}
    = \begin{bmatrix}
    A & E \\
    0_{3 \times 3} & I_{3\times3}
    \end{bmatrix} \begin{bmatrix}
    x \\
    c
    \end{bmatrix} + 
    \begin{bmatrix}
    B \\
    0_{3\times3}
    \end{bmatrix} u
    +\begin{bmatrix}
    L_x \\
    L_c
    \end{bmatrix}
    (y - \hat{y})
\end{equation}

where
$x=[p^\top,v^\top]^\top\in\mathbb{R}^6$ contains position and velocity,
$u\in\mathbb{R}^3$ contains roll, pitch, and total-thrust commands, and
$A$ and $B$ are obtained by linearizing the quadrotor dynamics about hover.
The matrix $E = \begin{bmatrix} 0_{3\times3} & (dt)I_{3\times3} \end{bmatrix}$ maps the estimated disturbance into the translational dynamics, where $dt$ is the timestep of the discrete dynamics. The disturbance is assumed constant over each prediction step, while $L_x$ and $L_c$ determine the state and disturbance corrections from the motion-capture position innovation $y-\hat{y}$.

The estimated state and disturbance are provided to the trajectory-optimization QP, which is solved at $45 \ \mathrm{Hz}$. Its output is tracked by the Crazyflie's onboard low-level controller running at $500 \ \mathrm{Hz}$. For the learned QP policy, the cost and disturbance-estimation parameters are optimized through differentiable closed-loop rollouts. The training loss is dominated by position-tracking error relative to the target. Training is performed using disturbances with randomized direction and magnitude, exposing the policy to a range of wind conditions.

We compare the learned QP policy against two manually tuned MPC baselines using Bryson's rule \cite{okyere2019lqr}. For both baselines, the disturbance observer gain is fixed to  $L_c = [1,1,1]^\top$, while the QP policy disturbance observer gain is learned. The first controller is tuned for an allowable position deviation of $10 \ \text{cm}$ in each direction, while the second uses a more aggressive value of $5 \ \text{cm}$. Increasing the position penalty further resulted in aggressive control and unreliable flight, providing a practical limit to manual tuning for this experiment. 

\begin{figure}[t]
    \centering
    \includegraphics[width=\columnwidth]{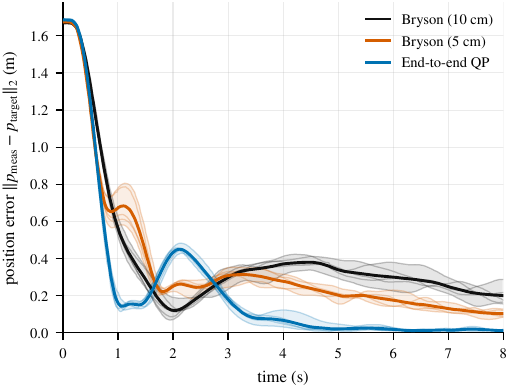}
  \caption{Closed-loop position error under crosswind for the learned end-to-end QP policy and two manually tuned MPC baselines based on Bryson’s rule. Thick curves show the mean over five trials, thin transparent curves show individual trials, and light shading marks the trial range.}
    \label{fig:wind_experiment_data}
\end{figure}

The results are shown in Fig.~\ref{fig:wind_experiment_data}. Thick curves show the mean position error over five trials, thin transparent curves show individual trials, and light shading marks their range. The learned QP policy converges substantially faster than both manually tuned baselines and drives the steady-state position error to zero without manual cost tuning. The baseline controllers are also able to reject the wind disturbance through their disturbance observers, but converge more slowly and retain a residual position error throughout the flight. Overall, the learned policy achieves both faster disturbance rejection and improved steady-state regulation.  




The hardware experiment shows that jointly tuning the structured controller and disturbance estimator from randomized closed-loop rollouts can substantially improve disturbance rejection compared with manual cost tuning. Together with the preceding experiments, this shows that the same QP policy abstraction can span nonlinear policy learning, robustness through domain randomization, and joint estimator-controller tuning on hardware.







\section{CONCLUSIONS}
\label{sec:conclusions}
We presented an end-to-end QP-based policy framework in which the optimization problem itself serves as a learnable policy representation. By separating the QP structure from the outer closed-loop objective, the framework provides a common perspective on robust control, neural-network policies, and domain-randomized robot learning while retaining the structure and interpretability of optimization-based control. Across simulation and hardware experiments, we showed that this abstraction can recover classical robustness objectives, learn nonlinear feedback policies, improve robustness to model mismatch, and jointly tune estimation and control under strong disturbances. 

\section*{ACKNOWLEDGMENT}
This material is based upon work supported by the National Science Foundation under Grand No. DGE2140739. 

\printbibliography{}

@inproceedings{adabag2026differentiable,
  title={Differentiable model predictive control on the gpu},
  author={Adabag, Emre and Greiff, Marcus and Subosits, John and Lew, Thomas},
  booktitle={International Conference on Learning Representations},
  volume={2026},
  pages={132305--132323},
  year={2026}
}

@inproceedings{romero2024actor,
  title={Actor-critic model predictive control},
  author={Romero, Angel and Song, Yunlong and Scaramuzza, Davide},
  booktitle={2024 IEEE International Conference on Robotics and Automation (ICRA)},
  pages={14777--14784},
  year={2024},
  organization={IEEE}
}

@article{reiter2026synthesis,
  title={Synthesis of model predictive control and reinforcement learning: Survey and classification},
  author={Reiter, Rudolf and Hoffmann, Jasper and Reinhardt, Dirk and Messerer, Florian and Baumg{\"a}rtner, Katrin and Sawant, Shambhuraj and Boedecker, Joschka and Diehl, Moritz and Gros, Sebastien},
  journal={Annual Reviews in Control},
  volume={61},
  pages={101045},
  year={2026},
  publisher={Elsevier}
}

@inproceedings{coulson2019data,
  title={Data-enabled predictive control: In the shallows of the DeePC},
  author={Coulson, Jeremy and Lygeros, John and D{\"o}rfler, Florian},
  booktitle={2019 18th European control conference (ECC)},
  pages={307--312},
  year={2019},
  organization={IEEE}
}

@article{markovsky2021behavioral,
  title={Behavioral systems theory in data-driven analysis, signal processing, and control},
  author={Markovsky, Ivan and D{\"o}rfler, Florian},
  journal={Annual Reviews in Control},
  volume={52},
  pages={42--64},
  year={2021},
  publisher={Elsevier}
}

@inproceedings{howell2021direct,
  title={Direct Policy Optimization using Deterministic Sampling and Collocation},
  author={Howell, Taylor A. and Fu, Chunjiang and Manchester, Zachary},
  booktitle={2021 IEEE International Conference on Robotics and Automation (ICRA)},
  year={2021},
  organization={IEEE}
}

@article{arrizabalaga2026differentiable,
  title={A Differentiable Interior-Point Method in Single Precision},
  author={Arrizabalaga, Jon and Tracy, Kevin and Manchester, Zachary},
  journal={arXiv preprint arXiv:2605.17913},
  year={2026}
}

@article{amos2019differentiable,
  title={Differentiable optimization-based modeling for machine learning},
  author={Amos, Brandon},
  year={2019}
}

@article{salzmann2023real,
  title={Real-time neural MPC: Deep learning model predictive control for quadrotors and agile robotic platforms},
  author={Salzmann, Tim and Kaufmann, Elia and Arrizabalaga, Jon and Pavone, Marco and Scaramuzza, Davide and Ryll, Markus},
  journal={IEEE Robotics and Automation Letters},
  volume={8},
  number={4},
  pages={2397--2404},
  year={2023},
  publisher={IEEE}
}

@article{xiao2023barriernet,
  title={Barriernet: Differentiable control barrier functions for learning of safe robot control},
  author={Xiao, Wei and Wang, Tsun-Hsuan and Hasani, Ramin and Chahine, Makram and Amini, Alexander and Li, Xiao and Rus, Daniela},
  journal={IEEE Transactions on Robotics},
  volume={39},
  number={3},
  pages={2289--2307},
  year={2023},
  publisher={IEEE}
}

@article{sambharya2026learning,
  title={Learning algorithm hyperparameters for fast parametric convex optimization},
  author={Sambharya, Rajiv and Stellato, Bartolomeo},
  journal={SIAM Journal on Mathematics of Data Science},
  volume={8},
  number={3},
  pages={649--676},
  year={2026},
  publisher={SIAM}
}

@inproceedings{blackmore2016autonomous,
  title={Autonomous precision landing of space rockets},
  author={Blackmore, Lars},
  booktitle={Frontiers of engineering: reports on leading-edge engineering from the 2016 symposium},
  volume={46},
  pages={15--20},
  year={2016},
  organization={The Bridge Washington, DC, USA}
}

@inproceedings{kuindersma2014efficiently,
  title={An efficiently solvable quadratic program for stabilizing dynamic locomotion},
  author={Kuindersma, Scott and Permenter, Frank and Tedrake, Russ},
  booktitle={2014 IEEE International Conference on Robotics and Automation (ICRA)},
  pages={2589--2594},
  year={2014},
  organization={IEEE}
}

@inproceedings{todorov2012mujoco,
  title={Mujoco: A physics engine for model-based control},
  author={Todorov, Emanuel and Erez, Tom and Tassa, Yuval},
  booktitle={2012 IEEE/RSJ international conference on intelligent robots and systems},
  pages={5026--5033},
  year={2012},
  organization={IEEE}
}

@article{hou2026world,
  title={World model for robot learning: A comprehensive survey},
  author={Hou, Bohan and Li, Gen and Jia, Jindou and An, Tuo and Guo, Xinying and Leng, Sicong and Geng, Haoran and Ze, Yanjie and Harada, Tatsuya and Torr, Philip and others},
  journal={arXiv preprint arXiv:2605.00080},
  year={2026}
}

@article{agrawal2019differentiating,
  title={Differentiating through a cone program},
  author={Agrawal, Akshay and Barratt, Shane and Boyd, Stephen and Busseti, Enzo and Moursi, Walaa M},
  journal={arXiv preprint arXiv:1904.09043},
  year={2019}
}

@inproceedings{amos2017optnet,
  title={Optnet: Differentiable optimization as a layer in neural networks},
  author={Amos, Brandon and Kolter, J Zico},
  booktitle={International conference on machine learning},
  pages={136--145},
  year={2017},
  organization={PMLR}
}

@article{agrawal2019differentiable,
  title={Differentiable convex optimization layers},
  author={Agrawal, Akshay and Amos, Brandon and Barratt, Shane and Boyd, Stephen and Diamond, Steven and Kolter, J Zico},
  journal={Advances in neural information processing systems},
  volume={32},
  year={2019}
}

@article{tracy2024differentiability,
  title={On the differentiability of the primal-dual interior-point method},
  author={Tracy, Kevin and Manchester, Zachary},
  journal={arXiv preprint arXiv:2406.11749},
  year={2024}
}

@book{krantz2002implicit,
  title={The implicit function theorem: history, theory, and applications},
  author={Krantz, Steven George and Parks, Harold R},
  volume={202},
  number={11},
  year={2002},
  publisher={Springer}
}

@article{bravo2026turbompc,
  title={TurboMPC: Fast, Scalable, and Differentiable Model Predictive Control on the GPU},
  author={Bravo-Palacios, Gabriel and Zhang, Jianghan and Pestrikov, Zachary and Plancher, Brian and Lew, Thomas},
  journal={arXiv preprint arXiv:2606.24039},
  year={2026}
}

@inproceedings{morton2025safe,
  title={Safe, task-consistent manipulation with operational space control barrier functions},
  author={Morton, Daniel and Pavone, Marco},
  booktitle={2025 IEEE/RSJ International Conference on Intelligent Robots and Systems (IROS)},
  pages={187--194},
  year={2025},
  organization={IEEE}
}

@article{holmes2025sdprlayers,
  title={Sdprlayers: Certifiable backpropagation through polynomial optimization problems in robotics},
  author={Holmes, Connor and D{\"u}mbgen, Frederike and Barfoot, Timothy D},
  journal={IEEE Transactions on Robotics},
  volume={41},
  pages={4120--4138},
  year={2025},
  publisher={IEEE}
}

@book{boyd1994linear,
  title={Linear matrix inequalities in system and control theory},
  author={Boyd, Stephen and El Ghaoui, Laurent and Feron, Eric and Balakrishnan, Venkataramanan},
  year={1994},
  publisher={SIAM}
}

@article{willems2005note,
  title={A note on persistency of excitation},
  author={Willems, Jan C and Rapisarda, Paolo and Markovsky, Ivan and De Moor, Bart LM},
  journal={Systems \& Control Letters},
  volume={54},
  number={4},
  pages={325--329},
  year={2005},
  publisher={Elsevier}
}

@inproceedings{basar1989dynamic,
  title={A dynamic games approach to controller design: Disturbance rejection in discrete time},
  author={Basar, Tamer},
  booktitle={Proceedings of the 28th IEEE Conference on Decision and Control,},
  pages={407--414},
  year={1989},
  organization={IEEE}
}

@article{lian2021koopman,
  title={Koopman based data-driven predictive control},
  author={Lian, Yingzhao and Wang, Renzi and Jones, Colin N},
  journal={arXiv preprint arXiv:2102.05122},
  year={2021}
}

@article{naf2025choose,
  title={Choose wisely: Data-driven predictive control for nonlinear systems using online data selection},
  author={N{\"a}f, Joshua and Moffat, Keith and Eising, Jaap and D{\"o}rfler, Florian},
  journal={arXiv preprint arXiv:2503.18845},
  year={2025}
}

@article{huang2023robust,
  title={Robust and kernelized data-enabled predictive control for nonlinear systems},
  author={Huang, Linbin and Lygeros, John and D{\"o}rfler, Florian},
  journal={IEEE Transactions on Control Systems Technology},
  volume={32},
  number={2},
  pages={611--624},
  year={2023},
  publisher={IEEE}
}

@misc{megretski20006,
  title={6.245: MULTIVARIABLE CONTROL SYSTEMS by},
  author={Megretski, Alexandre},
  year={2000},
  publisher={Department of Electronic Engineering and Computer Science, MIT}
}

@article{dietz2008analysis,
  title={Analysis and control of uncertain systems by using robust semi-definite programming},
  author={Dietz, Sjoerd Gerard},
  journal={Delft University of Technology, The Netherlands},
  year={2008}
}

@book{zhou1996robust,
  title={Robust and optimal control},
  author={Zhou, Kemin and Doyle, John Comstock and Glover, Keith and others},
  volume={40},
  year={1996},
  publisher={Prentice hall New Jersey}
}

@article{guth2025approximation,
  title={Approximation of risk-averse optimal feedback control},
  author={Guth, Philipp A and Kunisch, Karl},
  journal={arXiv preprint arXiv:2508.15618},
  year={2025}
}

@article{okyere2019lqr,
  title={LQR controller design for quad-rotor helicopters},
  author={Okyere, Emmanuel and Bousbaine, Amar and Poyi, Gwangtim T and Joseph, Ajay K and Andrade, Jose M},
  journal={The Journal of Engineering},
  volume={2019},
  number={17},
  pages={4003--4007},
  year={2019},
  publisher={Wiley Online Library}
}

@article{luenberger1966observers,
  title={Observers for multivariable systems},
  author={Luenberger, David},
  journal={IEEE transactions on automatic control},
  volume={11},
  number={2},
  pages={190--197},
  year={1966},
  publisher={IEEE}
}

@inproceedings{pannocchia2015offset,
  title={Offset-free tracking MPC: A tutorial review and comparison of different formulations},
  author={Pannocchia, Gabriele},
  booktitle={2015 European control conference (ECC)},
  pages={527--532},
  year={2015},
  organization={IEEE}
}
\end{document}